\documentclass[conference, letterpaper]{IEEEtran}
\IEEEoverridecommandlockouts
\renewcommand{\IEEEkeywordsname}{Keywords}
\usepackage[switch]{lineno}
\usepackage{amsmath,amsfonts}
\usepackage{algorithmic}
\usepackage{algorithm}
\usepackage{array}
\usepackage[caption=false,font=footnotesize,labelfont=sf,textfont=sf]{subfig}
\usepackage{textcomp}
\usepackage{stfloats}
\usepackage{url}
\usepackage{verbatim}
\usepackage{graphicx}
\usepackage{cite}
\usepackage{url, tabularx, booktabs}
\newcolumntype{P}[1]{>{\centering\arraybackslash}m{#1}}

\usepackage{tikz}

\newcommand\copyrighttext{%
  \footnotesize \textcopyright 2026 IEEE. Personal use of this material is permitted.
  Permission from IEEE must be obtained for all other uses, in any current or future
  media, including reprinting/republishing this material for advertising or promotional
  purposes, creating new collective works, for resale or redistribution to servers or
  lists, or reuse of any copyrighted component of this work in other works.}
\newcommand\copyrightnotice{%
\begin{tikzpicture}[remember picture,overlay]
\node[anchor=south,yshift=10pt] at (current page.south) 
  {\fbox{\parbox{\dimexpr\textwidth-\fboxsep-\fboxrule\relax}{\copyrighttext}}};
\end{tikzpicture}%
}

\def\BibTeX{{\rm B\kern-.05em{\sc i\kern-.025em b}\kern-.08em
    T\kern-.1667em\lower.7ex\hbox{E}\kern-.125emX}}
\begin{document}
\bstctlcite{IEEEexample:BSTcontrol}
\title{TASTE: Throughput-Aware Batch Size Tuning
for \\ On-Device Edge Learning 
\thanks{This research was funded by the German Federal Ministry of Research, Technology, and Space (BMFTR), and the EU ChipsJU as part of the project RIGOLETTO under Grant 16MEE0547 and 101194371, respectively.}
}

\author{\IEEEauthorblockN{Avik Bhatnagar, Federico Nicolás Peccia, Oliver Bringmann}
\IEEEauthorblockA{
\textit{FZI Research Center for Information Technology, University of Tübingen} \\
Germany \\
bhatnagar@fzi.de, peccia@fzi.de, oliver.bringmann@uni-tuebingen.de}
}




\maketitle
\copyrightnotice
\begin{abstract}
The rise of privacy-preserving artificial intelligence (AI) has shifted the focus of model adaptation and personalization towards on-device learning, where deep learning models are fine-tuned directly on edge hardware using local user data. However, this shift requires optimization of deep learning training on resource-constrained hardware to maximize throughput while maintaining predictive accuracy. This paper introduces a novel technique for on-device model training that incorporates an efficient Bayesian optimization-based batch size tuning approach to maximize hardware throughput. To evaluate the impact of this hyperparameter on the learning dynamics, we investigated two distinct paradigms: standard supervised learning (SL) and online continual learning (CL). Experimental results across various edge devices demonstrate a throughput ceiling, beyond which increasing the batch size yields no additional throughput gains. The proposed tuning approach identifies the optimal batch size, which, when combined with gradient accumulation and linear learning rate scaling, achieves up to a 2X increase in training throughput on platforms such as Raspberry Pi 4 compared to maximum batch sizes, without compromising model accuracy. Furthermore, in the CL paradigm, we demonstrate that optimal batch sizes maintain the stability-plasticity balance required for incremental learning, effectively mitigating catastrophic forgetting while maximizing computational efficiency on edge-hardware. 
\end{abstract}

\begin{IEEEkeywords}
On-device learning, Throughput optimization, Resource-constrained hardware, Continual learning, Supervised learning, Gradient accumulation, Learning rate scaling
\end{IEEEkeywords}

\section{Introduction}
\label{sec:Introduction}

The deployment of AI models in real-world environments frequently encounters the challenge of data distribution shifts, in which the operating conditions and environments can deviate significantly from the original training distribution. Such shifts are prevalent across speech recognition (acoustic variations, noise), computer vision (varying sensors, lighting, weather, sim-to-real gaps), and specialized fields like healthcare or activity monitoring (physiological traits, movement speeds). Overcoming these shifts requires robust adaptation strategies; to address this, on-device learning has emerged as an optimal solution. By facilitating model adaptation directly on edge hardware, this approach eliminates the need to transmit data to centralized cloud servers. Consequently, on-device learning reduces infrastructure and communication overhead, while enhancing data privacy by locally maintaining sensitive information. In addition, on-device learning enables highly personalized on-device AI experiences tailored to specific user contexts. For example, local learning capability has proven beneficial for enhancing the performance of keyword spotting systems for voice commands \cite{Rusci_2025}, achieving robust human pose estimation for nanodrones \cite{cereda2024ondeviceselfsupervisedlearningvisual}, enhancing battery power predictions in real-world electric vehicle scenarios \cite{11344518}, and achieving accurate human activity recognition through adaptation to user biometric data \cite{kang2024ondevicetrainingempoweredtransfer}. Within these applications, the learning process typically follows one of two primary paradigms: offline supervised learning, in which the model is fine-tuned on a locally stored dataset tailored to a specific user, or online continual learning, which enables the model to incrementally incorporate new data and tasks without losing previously acquired knowledge.


Regardless of the selected learning paradigm, batch size remains a critical hyperparameter that dictates both the memory requirements and computational overhead required for model training. On a server-class cloud infrastructure, it is believed that larger batch sizes lead to better resource utilization and faster convergence, primarily because of the highly parallelized execution capabilities of modern GPU architectures. However, while maximizing the batch size can increase hardware throughput, it can also significantly affect the learning performance. Larger batch sizes minimize the gradient variance, providing a more accurate gradient representation across the dataset and resulting in stable and faster convergence \cite{goyal2018accuratelargeminibatchsgd}. Conversely, the existing literature also indicates that excessively large batch sizes may lead to a generalization gap in which the model converges toward sharper minima \cite{keskar2017largebatchtrainingdeeplearning}. In contrast, smaller batch sizes introduce stochastic gradient noise, which encourages convergence toward flatter minima and often results in a more robust and well-generalized solution \cite{keskar2017largebatchtrainingdeeplearning}.

Existing research has extensively explored methods for maximizing hardware utilization and throughput for deep neural network (DNN) workloads. One such method is DyNet \cite{neubig2017dynetdynamicneuralnetwork}, a foundational work that supports dynamic batching in natural language processing. DyNet unifies model declaration with on-the-fly computational graph construction for every new input. Although this strategy allows highly flexible network structures, it requires continuous resource utilization to rebuild the graph for each instance. In addition, such dynamic declarations miss global optimizations that can be performed on static computational graphs. There are also high data movement costs, as it relies on explicit memory gather/scatter operations, as demonstrated in the ED-Batch work \cite{chen2023edbatchefficientautomaticbatching}. ED-Batch utilizes finite state machines (FSMs) integrated with reinforcement learning (RL) to find specialized batching policies for DNN workloads. Additionally, it includes a policy-aware tree-based algorithm for memory planning that reduces data movement overheads. By utilizing a predefined static graph, ED-Batch substantially outperforms DyNet by minimizing graph construction and runtime latencies. More recently, ACROBAT \cite{fegade2024acrobatoptimizingautobatchingdynamic}, built upon the TVM framework \cite{chen2018tvmautomatedendtoendoptimizing}, employs a hybrid static (compiler) and dynamic (runtime) analysis to generate optimized tensor kernels for auto-batching dynamic deep learning workloads. However, because the framework was developed for Nvidia GPUs, the optimizations are highly hardware-specific, such as reducing CUDA API runtime overheads.

A critical limitation of these existing works is their primary focus on server-class hardware and inference-only DNN workloads. The recurring computational costs associated with graph construction, data movement overheads, and the complexity of the underlying optimization algorithms (such as reinforcement learning or tree-based planning) make these approaches less suitable for resource-constrained edge-CPU devices. Crucially, unlike inference batching, training batching must account for intermediate activation storage required for backpropagation. In such environments, a static, pre-compiled alternative that accounts for strict memory and energy constraints during throughput optimization is often more viable.

Therefore, it is critical to investigate the inherent relationship between batch size and maximum achievable hardware performance and its direct implications on model accuracy during training. In particular, this is relevant to resource-limited embedded hardware frequently utilized in edge computing environments. In these environments, limited processing power is not the only hurdle; unlike cloud- or server-based training, edge devices cannot support the storage of massive datasets required for large-batch training. Furthermore, high-bandwidth memory architectures and rapid data transfer mechanisms, as observed in high-end GPUs, are absent in embedded systems.

We introduce \textbf{T}hroughput-\textbf{A}ware Batch \textbf{S}ize \textbf{T}uning for On-Device \textbf{E}dge Learning (TASTE). The TASTE technique is designed to achieve two primary objectives: (i) a systematic approach to determining an optimal batch size that maximizes computational throughput, which is not necessarily the maximum size supported by the hardware, and (ii) adaptively adjusting the training regimes to prevent the degradation in learning performance that can occur when throughput-optimized configurations replace initial batch size implementations. This dual approach is particularly advantageous for edge devices deployed in dynamic real-world environments because it enables accelerated training cycles without significant accuracy loss that can arise from differences between the original cloud-based training parameters and the proposed max-throughput on-device configuration. The main contributions of this paper are summarized as follows:


\begin{itemize}
\item We propose an on-device model-training technique with a Bayesian optimization-based batch size tuning method designed to maximize the training throughput on resource-constrained hardware.
\item We identified a throughput ceiling across diverse edge-CPUs and determined the optimal batch sizes, resulting in significant speedups of up to $2X$ in training throughput compared with the standard practice of using the maximum possible batch size.
\item We validate the approach via image classification across two distinct training paradigms: offline supervised and online continual learning, demonstrating that with tailored training regimes, the optimized max-throughput batch sizes achieve learning performance comparable to reference implementations.
\end{itemize}

To our knowledge, this study is the first to present the feasibility to maximize training throughput on resource-constrained edge-CPU devices, complemented by appropriate changes to training regimes for two distinct learning paradigms. This methodology ensures efficient on-device training without compromising learning performance in an image classification task. The demonstrated methodology can also be seamlessly transferred to various other combinations of deep learning models, applications, and specialized hardware devices.

The remainder of this paper is organized as follows. Section \ref{sec:Methodology} discusses the methodology for enabling on-device model training on resource-constrained hardware, the proposed batch size tuning approach for throughput optimization, and the evaluation of training performance. Sections \ref{sec:Experimental Setup} and \ref{sec:Results} detail the experimental setup and discuss results, respectively. Finally, Section \ref{sec:Conclusion} presents conclusions and outlines future work.
\section{Methodology}
\label{sec:Methodology}

\subsection{On-device Model Training}

To enable on-device model training, the proposed technique supports the execution of both the inference forward pass to calculate the model output and a subsequent backward pass to compute the loss and model parameter gradients. We utilized the ONNX training library \cite{onnxruntime} to generate the training artifact for the target deep learning model. The library extends the original inference-only forward computation graph by appending the required gradients and loss operators for the backward pass. This allows the model to update its parameters and adapt iteratively to the local input data. Notably, these gradient operations are composed of standard neural network operators already optimized for model inference. For instance, the gradients for a 2D convolutional layer in a Convolutional Neural Network (CNN) are derived as follows:

\begin{equation}
Y = X \ast W + B = \text{conv2d}\left(X, W\right) + B\label{eq1}
\end{equation}
\begin{equation}
\frac{\partial L}{\partial W} = X \ast \frac{\partial L}{\partial Y} = \text{conv2d}\left(X, \frac{\partial L}{\partial Y}\right)\label{eq2}
\end{equation}
\begin{equation}
\begin{split}
\frac{\partial L}{\partial X} &= \frac{\partial L}{\partial Y} \ast_{full} W_{rot180} \\ 
&= \text{conv2d\_transpose}\left(\frac{\partial L}{\partial Y}, W\right)
\end{split}
\label{eq3}
\end{equation}
\begin{equation}
\frac{\partial L}{\partial B} = \text{reduce\_sum}\left(\frac{\partial L}{\partial Y}\right)\label{eq4}
\end{equation}

During the forward pass, for a given input $X$, weight kernel $W$, and bias $B$, the output $Y$ of the convolution layer is computed using \eqref{eq1}. In the subsequent backward pass, the gradients of the loss $L$ with respect to the weight, input, and bias are computed using the upstream gradient $\frac{\partial L}{\partial Y}$ (i.e., the gradient of the loss with respect to the output) to update the layer's parameters. As defined in \eqref{eq2}, the weight gradient is obtained when the input is convolved with this output gradient. Similarly, the input gradient can be calculated by convolving the output gradient with a {$180$\textdegree} rotated kernel $W$ using full padding, which can also be expressed as a transposed convolution, as shown in \eqref{eq3}. Finally, \eqref{eq4} demonstrates that the bias gradient is calculated by summing the output gradient across the batch, height, and width dimensions for every channel. As illustrated in these equations, gradient calculations can be expressed entirely in terms of standard convolution, transposed convolution, and reduced sum operations.

For model deployment, the approach leverages the Apache TVM \cite{chen2018tvmautomatedendtoendoptimizing} compiler and optimization library. We established the on-device training capabilities by implementing support for gradient operators for the convolutional and fully connected layers of the input ONNX CNN model. To minimize memory overhead and computational latency, TVM-driven optimizations are utilized, including operator fusion, constant folding, and layout transformations. These optimizations result in a static computational graph that includes both forward and backward passes and is deployable across various hardware backends, including RISC-V and ARM cores supported by the library. Additionally, TVM offers hardware-specific optimizations for efficient execution on the available computing architecture. These include loop transformations and tensorizations, for example, through vectorization for devices supporting Arm Neon or the RISC-V Vector (RVV) extension \cite{peccia2025tensorprogramoptimizationriscv}. Finally, deployment to resource-constrained hardware is facilitated by TVM's Remote Procedure Call (RPC) based infrastructure, which enables the transmission of the compiled binary from the host device to the target edge platform for on-device execution and performance evaluation.

\subsection{Batch Size Tuning for Training Throughput Optimization}
With the established on-device model training, the batch size can be systematically tuned to maximize the training throughput. The maximum supported batch size ($bs_{\text{max}}$) is constrained by the memory available on a specific embedded device. Consequently, the search space $S$ for the optimal batch size is defined as $S = \{1, 2, 4, 6, \dots, bs_{\text{max}}\}$, following the common practice of evaluating even-numbered increments. Various hyperparameter tuning strategies can be employed to determine the optimal batch size. While a traditional Grid-Search ensures a global optimum by exhaustively evaluating every candidate in $S$, its computational overhead becomes prohibitive as the hardware's memory ($bs_{\text{max}}$) increases. Conversely, Random Search offers a stochastic alternative that improves efficiency but often fails to guarantee convergence within a limited number of iterations and lacks the heuristic guidance required for rapid convergence. Furthermore, simpler heuristics, such as profiling strictly by powers of two, risk missing optimal non-power-of-two even-numbered peaks.

\begin{figure}[htbp]
    \centering
    \includegraphics[width=\columnwidth]{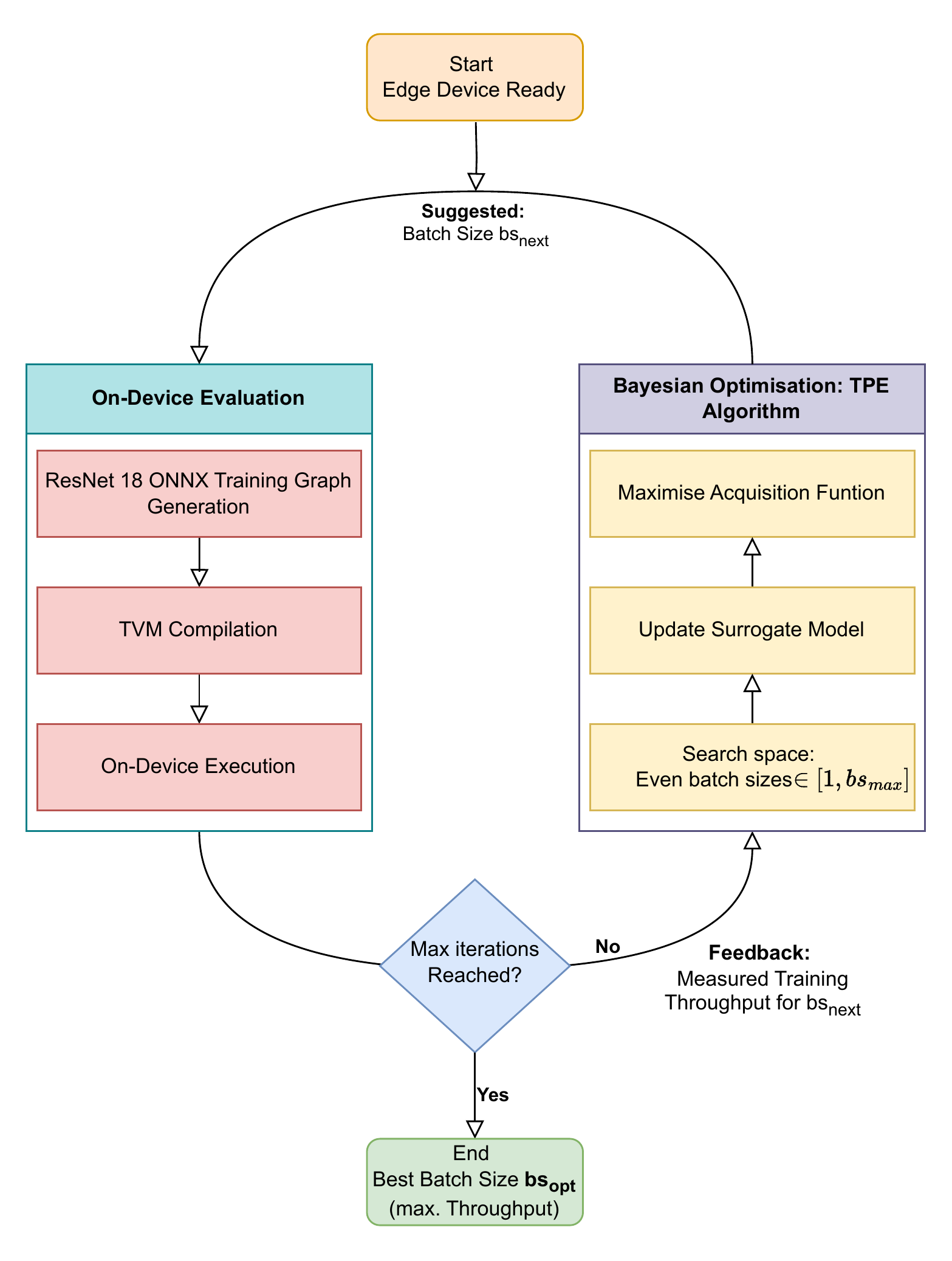} \\[-2ex]
    \caption{Overview of the proposed on-device training technique with Bayesian optimization-based batch size tuning for throughput optimization. The workflow illustrates the iterative process of on-device measurement and batch-size selection to identify the optimal configuration, $bs_{\text{opt}}$.}
    \label{fig: Data_dist}
\end{figure}

To achieve a more efficient yet granular search, we leveraged the hyperopt library \cite{bergstra2012makingsciencemodelsearch} for Bayesian Optimization utilizing the Tree-structured Parzen Estimator (TPE) algorithm \cite{NIPS2011_86e8f7ab}. By treating throughput maximization as an objective function, the TPE algorithm constructs a probabilistic model to intelligently select the next candidate batch size ($bs_{\text{next}}$). This allows a transition from the initial exploration phase to the exploitation phase, targeting the region of the search space that yields the highest performance. To ensure a constrained search time, the number of search iterations was limited to half the size of the search space. Consequently, our approach achieves a twofold reduction in search iterations compared with the exhaustive Grid Search while consistently identifying near-optimal configurations ($bs_{\text{opt}}$) that maximize throughput. Crucially, evaluating each candidate requires generating a batch-size-specific ONNX training graph, model compilation, device deployment, and benchmark runs. Ultimately, this iteration reduction yields significant potential cumulative time and energy savings when scaled across fleets of edge devices undergoing continuous on-device training. The resulting optimization workflow is illustrated in Fig. \ref{fig: Data_dist}, ensuring that continuous learning edge systems maintain peak throughput under varying resource constraints.

\begin{figure*}
    \centering
    \setlength{\tabcolsep}{2pt} 
    \begin{tabular}{cc}
        \includegraphics[width=0.49\textwidth]{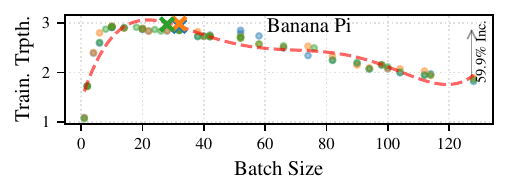} & 
        \includegraphics[width=0.49\textwidth]{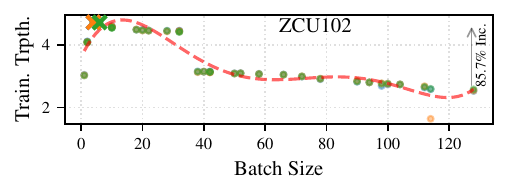} \\[-2ex] 
        \includegraphics[width=0.49\textwidth]{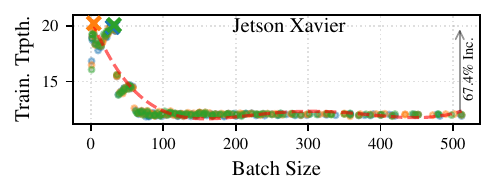} & 
        \includegraphics[width=0.49\textwidth]{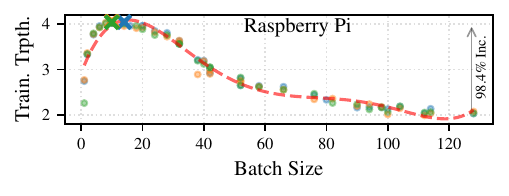} \\[-2ex] 
    \end{tabular}
    \includegraphics[width=\textwidth]{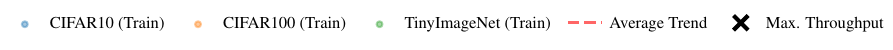} \\[-3ex] 
    \caption{The training throughput [imgs/sec] observed across the selected edge devices as a function of batch size. The crosses denote the identified optimal batch size ($bs_{\text{opt}}$) for each platform and dataset. The plot also highlights the mean throughput improvement relative to the $bs_{\text{max}}$ across all datasets.}
    \label{fig:train_trpt}
\end{figure*}

\subsection{Training Performance Evaluation}

The proposed on-device model adaptation was evaluated across two distinct training paradigms for image classification. The first paradigm employs the offline standard supervised learning (SL), in which the model is trained on a static dataset encompassing all target classification classes. In this setting, the complete dataset is available at the start of the training, and the data distribution is assumed to be independent and identically distributed (i.i.d.) until the model converges across multiple epochs. In contrast, the second paradigm is continual learning (CL), which requires the model to adapt dynamically to a non-stationary data stream whose distribution evolves over time. Unlike the supervised setting, there is no prior access to the entire dataset; rather, the model must learn incrementally from new classes as they become available. This sequential learning process introduces significant complexity, primarily because of catastrophic forgetting, that is, the phenomenon in which the model’s performance on previously learned tasks degrades as it adapts to novel data distributions.

The CL paradigm is further categorized into \textit{task-incremental learning} (TIL) and \textit{class-incremental learning} (CIL) settings. In the TIL setting, the training and evaluation phases comprise distinct, sequential tasks, each accompanied by a specific task identifier. These Task-IDs indicate the relevant subset of classes, allowing the model to isolate the training and inference to task-specific classification heads. Conversely, the CIL setting presents a more rigorous challenge because the model must adapt to new classes over time without the assistance of the classification boundaries provided by the task IDs during inference. Consequently, the model must maintain a unified classifier that can incrementally incorporate novel classes while simultaneously preserving the classification accuracy of the previously encountered classes.

Various CL strategies have been developed to mitigate catastrophic forgetting. 
Experience Replay (ER) \cite{chaudhry2019tinyepisodicmemoriescontinual} serves as a robust baseline. By leveraging a tiny episodic memory, ER is shown to outperform non-replay methods, including basic fine-tuning and regularization-based Elastic Weight Consolidation (EWC) technique \cite{Kirkpatrick_2017}, as well as more sophisticated Architecture-based techniques such as Dynamically Expandable Networks (DEN) \cite{yoon2018lifelonglearningdynamicallyexpandable} and rehearsal strategies such as Averaged Gradient Episodic Memory (A-GEM) \cite{chaudhry2019efficientlifelonglearningagem} and Meta-Experience Replay (MER) \cite{riemer2019learninglearnforgettingmaximizing}. Given its superior performance with only a marginal increase in computational overhead, which can be limited by restricting the size of memory buffers, ER has emerged as a particularly effective and lightweight solution for resource-constrained on-device CL environments \cite{10137046,math13142257,Ravaglia_2021}.

Prior research indicates that the selection of training hyperparameters, specifically the learning rate, batch size, and dropout regularization, has a significant influence on both the classification accuracy and the mitigation of catastrophic forgetting in CL settings \cite{mirzadeh2020understandingroletrainingregimes}. Similarly, within the SL paradigm, studies have suggested that larger batch sizes yield more stable and accurate gradient estimations. Furthermore, the ratio of batch size to learning rate has been shown to directly impact the generalization capabilities of the trained model \cite{NEURIPS2019_dc6a7071}. 

\begin{equation}
\eta_{opt} = \eta_{base} \times \frac{bs_{opt}}{bs_{base}}\label{eq5}
\end{equation}

Acknowledging the interdependence of these variables, this study also adjusts the training regime, along with our proposed approach, to use an optimized batch size that maximizes throughput without compromising performance. In the CL paradigm, optimal learning performance was observed when the obtained batch size was paired with a linearly scaled learning rate, $ \eta_{opt} $, as defined in Eq. \eqref{eq5}. This scaling was performed relative to the base learning rate configurations typically used in standard benchmarks. To achieve the benefits of larger batch sizes in SL on resource-constrained hardware, we employed gradient accumulation. This approach involves partitioning a large effective batch into smaller mini-batches of size $bs_{\text{opt}}$ and accumulating the resulting gradient. A single weight update is performed only after all constituent mini-batches are processed, thereby enabling model updates similar to those of large-batch training without exceeding the memory limits of the edge hardware.

\begin{figure*}
    \centering
    \setlength{\tabcolsep}{2pt} 
    \begin{tabular}{cc}
        \includegraphics[width=0.49\textwidth]{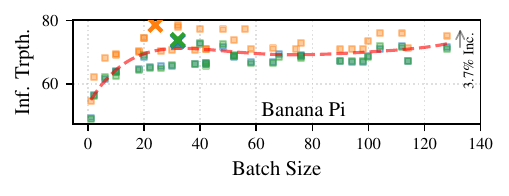} & 
        \includegraphics[width=0.49\textwidth]{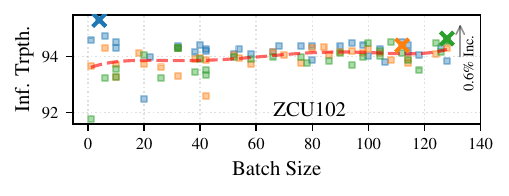} \\[-2ex] 
        \includegraphics[width=0.49\textwidth]{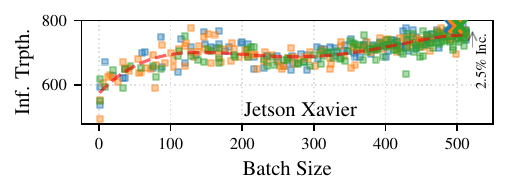} & 
        \includegraphics[width=0.49\textwidth]{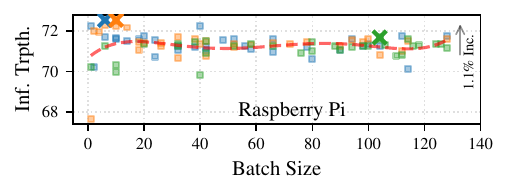} \\[-2ex] 
    \end{tabular}
    \includegraphics[width=\textwidth]{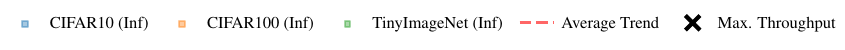} \\[-3ex] 
    \caption{The inference throughput [imgs/sec] observed across the selected edge devices as a function of batch size. The crosses denote the identified optimal batch size ($bs_{\text{opt}}$) for each platform and dataset. The plot also highlights the mean throughput improvement relative to the $bs_{\text{max}}$ across all datasets.}
    \label{fig:inf_trpt}
\end{figure*}

\section{Experimental Setup}
\label{sec:Experimental Setup}

The proposed approach was evaluated on the CIFAR-10, CIFAR-100, and TinyImageNet datasets, comprising 10, 100, and 200 classification classes, respectively. The ResNet-18 architecture was selected as the primary backbone for evaluation, consistent with the established benchmarks in CL \cite{chaudhry2019tinyepisodicmemoriescontinual}, \cite{chaudhry2019efficientlifelonglearningagem}. Experiments were conducted across distinct edge devices with varying computational and memory profiles, as detailed in Table \ref{tab1}. These devices feature multiple processor cores that support both ARM and RISC-V ISAs. The maximum supported batch size ($bs_{\text{max}} \in \{128, 512\}$) for each platform was determined based on the available hardware memory. To ensure a homogeneous comparison and isolate the performance of general-purpose floating-point processing, the execution was restricted to the CPU across all platforms, including the NVIDIA Jetson Xavier, thereby excluding the influence of specialized hardware accelerators. However, the proposed technique remains extensible to specialized accelerators and GPUs integrated in TVM for accelerated on-device learning applications. Notably, for the Banana Pi and ARM platforms, the implementation leverages RVV and Neon instructions by enabling vectorization with 256 and 128 bit vector lengths, respectively.

\begin{table}[htbp]
\caption{Hardware Specifications of Evaluated Edge-CPUs}
\begin{center}
\footnotesize
\begin{tabularx}{\columnwidth}{|X|P{0.18\linewidth}|P{0.14\linewidth}|P{0.16\linewidth}|P{0.14\linewidth}|}
\hline
\textbf{Properties} & \textbf{Banana Pi BPi-F3} & \textbf{Raspberry Pi 4B} & \textbf{NVIDIA Jetson Xavier} & \textbf{AMD ZCU102} \\
\hline
Compute & RISC-V SpacemiT K1 Octa-core & ARM A72 Quad-core & Carmel Octa-core & ARM A53 Quad-core \\
\hline
Clock Freq. & 1.6GHz & 1.8GHz & 2.2GHz & 1.2GHz \\
\hline
Memory & 4GB & 4GB & 16GB & 4GB \\
\hline
\end{tabularx}
\label{tab1}
\end{center}
\end{table}

The proposed on-device training technique is used to deploy the model onto edge devices capable of local loss and gradient computations. For the SL paradigm, we utilized a gradient accumulation strategy, where the derived optimal micro-batch size ($bs_{opt}$) was accumulated to achieve an effective global batch size of 128. The backbone architecture consisted of a pretrained PyTorch ResNet-18 model, originally trained on ImageNet, with the final classification layer modified to match the output dimensions of the target dataset. A constant learning rate of 0.1 was maintained across all batch sizes, and the models were trained for 200 epochs. 

The CL performance was benchmarked using the mammoth continual learning framework \cite{boschini2022class}, which was extended to include the same ResNet-18 backbone used in the SL experiments. The datasets are presented sequentially, providing classes with or without task identifiers to evaluate TIL and CIL, respectively. To address resource constraints on edge devices, an online CL setting was explored, where the model processes the training stream in a single pass over one epoch, thereby eliminating the overhead associated with continuously storing the entire input data stream. To benchmark the CL performance at the high-throughput batch size ($bs_{opt}$) against the reference and maximum batch sizes, we adopt the Experience Replay (ER) technique\cite{chaudhry2019tinyepisodicmemoriescontinual} utilizing a reservoir sampling strategy to maintain the replay memory buffer. Following this reference configuration, the base mini-batch was set to 10; thus, each training step utilized an equal distribution of 10 new and 10 buffered replay samples for a total of 20 samples. The base learning rate of 0.03 was scaled proportionally for various mini-batch sizes, including 2, 5, 7, 14, 16, 64, and 256, to align with the optimal and maximum batch sizes identified in this study. To minimize computational and memory overhead, the memory buffer size was fixed at 100 for the CIFAR datasets and 200 for TinyImageNet, effectively evaluating the performance with a constraint of one sample per class. However, for experiments involving a higher batch size of 512 (or mini-batch size of 256), the buffer size was adjusted to 300. In both training paradigms, Stochastic Gradient Descent (SGD) was employed as the optimizer for parameter updates. To ensure statistical robustness, the training performance was benchmarked across five randomized seeds.

\begin{figure*}[t] 
    \centering
    \begin{minipage}[b]{0.32\textwidth}
        \centering
        \includegraphics[width=\textwidth]{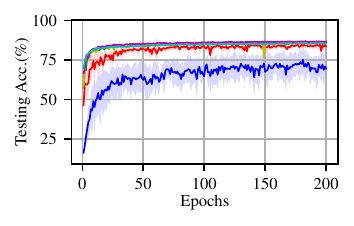}
        
        \hfill\includegraphics[width=0.85\textwidth]{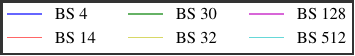}
    \end{minipage}%
    \hfill 
    \begin{minipage}[b]{0.32\textwidth}
        \centering
        \includegraphics[width=\textwidth]{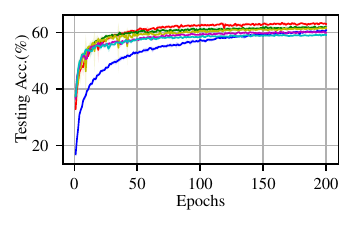}
        
        \hfill\includegraphics[width=0.85\textwidth]{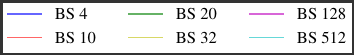}
    \end{minipage}%
    \hfill 
    \begin{minipage}[b]{0.32\textwidth}
        \centering
        \includegraphics[width=\textwidth]{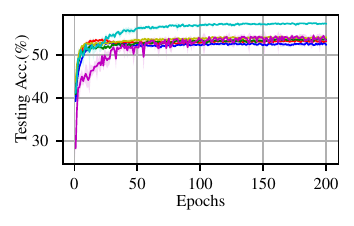}
        
        \hfill\includegraphics[width=0.85\textwidth]{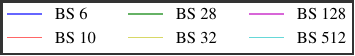}
    \end{minipage}%
    \vspace{-4pt}
    \caption{The average testing accuracies for standard supervised learning across CIFAR-10 (left), CIFAR-100 (middle), and TinyImageNet (right) datasets. The plots compare the performance of all the optimal batch sizes ($bs_{\text{opt}}$) against the $bs_{\text{max}} (128$ and $512$), utilizing gradient accumulation to maintain accuracy.}
    \label{fig:total_comparison_sl}
\end{figure*}
\section{Results and Discussion}
\label{sec:Results}


\subsection{Hardware Throughput Characterization and $bs_{\text{opt}}$ Identification}

The training and inference throughputs for the ResNet-18 model across multiple datasets are shown in Figs. \ref{fig:train_trpt} and \ref{fig:inf_trpt}, respectively. The depicted batch sizes were selected for evaluation using the proposed batch-size tuning approach. The inference throughput is defined as the number of images inferred per second, which is inversely proportional to the execution latency. Meanwhile, the training throughput accounts for the additional computational overhead and latency required for the loss and gradient calculations.

\begin{figure}[htbp]
     \centering
     \begin{minipage}{\columnwidth}
         \centering
         \hspace{1cm}\includegraphics[width=0.8\columnwidth]{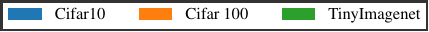}
     \end{minipage}
     \begin{minipage}{\columnwidth}
         \centering
         \includegraphics[width=\linewidth]{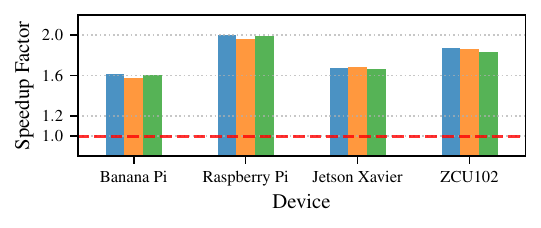} 
     \end{minipage}
     \vspace{-15pt}
     \caption{Observed training speedup observed with $bs_\text{opt}$ compared to $bs_\text{max}$ for CIFAR and TinyImageNet datasets across all edge devices.}
     \label{fig:overall_speedup_results}
\end{figure}

As indicated by the average trend line in Fig. \ref{fig:train_trpt}, a distinct pattern emerges across all edge devices where the training throughput peaks and subsequently declines as the batch size increases. This phenomenon represents a throughput ceiling, where hardware reaches its maximum processing rate for a given input batch size. In this study, this peak was identified as the optimal batch size ($bs_{\text{opt}}$) for training on resource-constrained hardware and is denoted by crosses in the plot. The average increase in training throughput across datasets relative to the maximum possible batch size is also highlighted in the plot. The bottleneck is due to the need to store intermediate activations from the ResNet-18 layers required for the backward pass, which scales with the batch size.


In contrast, inference does not require the retention of intermediate activations. Consequently, increasing the batch size is expected to improve the inference throughput, provided that the device has sufficient computational capacity. This trend is confirmed by the average inference throughput observed across the devices, as shown in Fig. \ref{fig:inf_trpt}. Although batch size tuning can be applied to model inference, the data suggest that larger batch sizes generally yield higher throughput. As shown in the plot, the optimal batch size for inference provides only a negligible average increase in throughput across the datasets compared with the maximum possible batch size. Thus, it can be concluded that for inference tasks, utilizing the maximum possible batch size is preferable, as the performance remains nearly identical to the baseline.

Although it may be intuitive to expect similar throughput values across the three datasets, given the shared ResNet-18 backbone, the observed optimal batch sizes for training and inference, marked with a cross in the plots, reveal slight variations. As anticipated, $bs_{\text{opt}}$ for inference generally remained closer to the maximum hardware limits. Fig. \ref{fig:overall_speedup_results} illustrates the observed training speedup, relative to the maximum possible batch size. The training throughput demonstrated a significant increase of at least $1.5\times$ across all devices, peaking at approximately $2\times$ on the Raspberry Pi platform. 

\begin{figure*}[t]
    \centering
    \subfloat[Class-Incremental Learning performance.\label{fig:class_il}]{%
        \centering
        \begin{minipage}[b]{0.32\textwidth}
            \centering 
            \includegraphics[width=\textwidth]{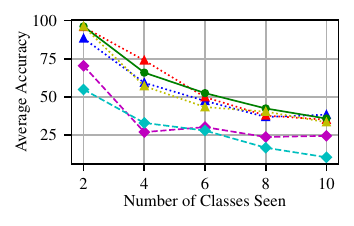}
             
            \hfill\includegraphics[width=0.85\textwidth]{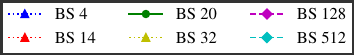}
        \end{minipage}%
        \hfill 
        \begin{minipage}[b]{0.32\textwidth}
            \centering
            \includegraphics[width=\textwidth]{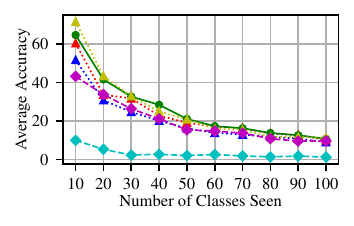}
            
            \hfill\includegraphics[width=0.85\textwidth]{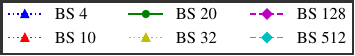}
        \end{minipage}%
        \hfill 
        \begin{minipage}[b]{0.32\textwidth}
            \centering
            \includegraphics[width=\textwidth]{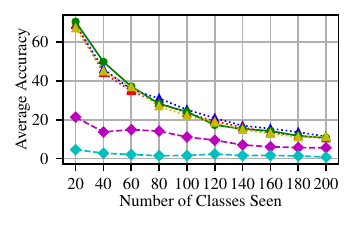}
            
            \hfill\includegraphics[width=0.85\textwidth]{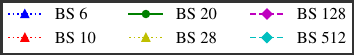}
        \end{minipage}%
    }
    
    \vspace{-6pt}

    \subfloat[Task-Incremental Learning performance.\label{fig:task_il}]{%
        \centering
        \begin{minipage}[b]{0.32\textwidth}
            \centering 
            \includegraphics[width=\textwidth]{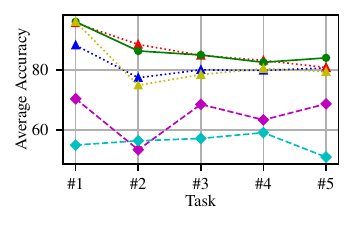}
             
            \hfill\includegraphics[width=0.85\textwidth]{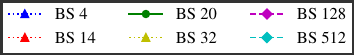}
        \end{minipage}%
        \hfill 
        \begin{minipage}[b]{0.32\textwidth}
            \centering
            \includegraphics[width=\textwidth]{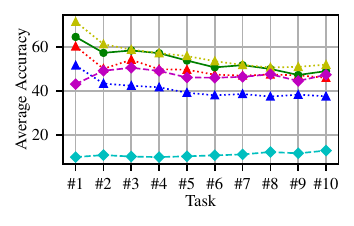}
            
            \hfill\includegraphics[width=0.85\textwidth]{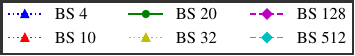}
        \end{minipage}%
        \hfill 
        \begin{minipage}[b]{0.32\textwidth}
            \centering
            \includegraphics[width=\textwidth]{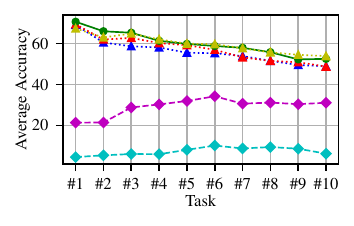}
            
            \hfill\includegraphics[width=0.85\textwidth]{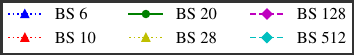}
        \end{minipage}%
    }
    \vspace{-2pt}
    \caption{Average testing accuracy as a function of the cumulative number of learned classes or tasks during online (a) CIL or (b) TIL across CIFAR-10 (left), CIFAR-100 (middle), and TinyImageNet (right) datasets. The plot compares accuracy trends across $bs_\text{opt}$ values (triangle markers), the reference batch size ($20$, solid green line), and $bs_\text{max}$ ($128$ and $512$, diamond markers), utilizing linear learning rate scaling to preserve learning performance.}
    \label{fig:total_comparison_cl_class}
\end{figure*}

\subsection{Impact of Optimized Batch Sizes on Training Performance}



A critical distinction arises during the training phase because of the varying complexities of the datasets. For classification tasks involving $10$, $100$, and $200$ classes, the training overhead for the identical ResNet backbone differed across the SL and CL paradigms. Fig. \ref{fig:total_comparison_sl} illustrates the standard supervised training performance using optimal batch sizes compared to the maximum possible batch sizes of $128$ and $512$. While larger batch sizes demonstrate robust performance, applying mini-batches and gradient accumulation techniques yields comparable testing accuracies.

For CIFAR-10, the simplest with $10$ classes, final test accuracy peaked at $86.76\%\pm0.11\%$ after $200$ epochs, though a pronounced performance degradation to $69.61\%\pm13.82\%$ occurred due to the unstable training at the smallest batch size of $4$. Interestingly, in the CIFAR-100 experiments, an optimal batch size of $10$ achieved $62.97\%\pm0.30\%$ accuracy, outperforming larger configurations like batch sizes $128$ ($59.91\%\pm0.46\%$) and $512$ ($59.08\%\pm0.31\%$). This phenomenon may be attributed to the stochastic noise introduced by smaller mini-batch sizes during the calculation of the Batch Normalization mean and variance. This noise effectively acts as a regularizer, improving generalization in the $100$-class problem. On the more challenging TinyImageNet dataset, performance scaled with larger batch sizes, rising from $52.37\%\pm0.30\%$ (batch size $6$) to a peak of $57.23\%\pm0.25\%$ (batch size $512$), as the $200$ classes can be better represented in each update.

In the Online CL paradigm, classes are presented to the model incrementally rather than simultaneously. The impact of the batch size on the class representation changes over time as the number of observed classes increases. Fig. \ref{fig:class_il} presents the training performance for CIL, where new classes are introduced in steps of 2, 10, and 20 for CIFAR-10, CIFAR-100, and TinyImageNet, respectively. The average test accuracy is plotted as a function of the cumulative number of classes learned by the model. Similarly, Fig. \ref{fig:task_il} depicts TIL performance, where classes are partitioned into 5 tasks for CIFAR-10 (2 classes per task) and 10 tasks for CIFAR-100 and TinyImageNet (10 and 20 classes per task, respectively). Following the experience replay-based CL implementation \cite{chaudhry2019tinyepisodicmemoriescontinual}, the baseline batch size was set to 20 (solid green line). 

In general, as the complexity of CL increases with the number of classes and tasks, average accuracy declines. However, in both incremental scenarios, the optimal batch sizes ($bs_{\text{opt}}$), coupled with a linearly scaled learning rate, achieved final-step accuracies within close margins of the baseline. E.g., final CIL accuracies for $bs_{\text{opt}}$ versus the baseline are $36.12\%\pm2.23\%$ ($bs_{\text{opt}} = 4$) vs. $34.48\%\pm1.21\%$ on CIFAR-10, $11.90\%\pm0.97\%$ ($bs_{\text{opt}}= 10$) vs. $10.98\%\pm0.60\%$ on CIFAR-100, and $11.57\%\pm0.30\%$ ($bs_{\text{opt}} = 6$) vs. $10.01\%\pm0.38\%$ on TinyImageNet; a similar trend holds for TIL. Crucially, these configurations outperform the maximum batch sizes \{$128$, $512$\}, which suffered absolute accuracy degradation of up to \{$11.67\%\pm3.14\%$, $21.01\%\pm3.67\%$\} in CIL and \{$15.02\%\pm11.72\%$, $28.54\%\pm3.81\%$\} in TIL on CIFAR-10, respectively, with similar degradation on other datasets. This divergence from standard SL trends is attributed to the stability-plasticity trade-off. Specifically, optimal batch sizes maintain the plasticity required to acquire new information while ensuring the stability necessary to mitigate catastrophic forgetting. Larger batch sizes result in fewer parameter updates for the same amount of data. Furthermore, given a fixed replay memory size, larger batches increase the frequency at which the same stored data points are reused, potentially leading to overfitting of the representative samples of older classes.


\section{Conclusion \& Future Work}
\label{sec:Conclusion}

This study demonstrated the potential to maximize training throughput across edge-CPU devices by identifying the optimal batch size without sacrificing learning performance or prediction accuracy for the image classification task. We proposed TASTE, an on-device training and deployment technique along with a Bayesian-optimization-based batch-size tuning. Our findings revealed that, unlike inference tasks, increasing the training batch size beyond a hardware-defined throughput ceiling yields no further performance gains because of memory bottlenecks associated with storing intermediate activations during the backward pass. The experimental results indicated that using the optimal batch size achieved a significant throughput speedup of at least $1.5\times$ across all evaluated edge devices, with a peak performance gain of approximately $2\times$ on the Raspberry Pi platform compared to the maximum batch size configuration. Furthermore, evaluations across supervised and online continual learning paradigms confirmed that by incorporating gradient accumulation and linear learning rate scaling, the identified optimal batch sizes achieved accuracies comparable to those of the reference implementations.


While this study evaluates the CNN-based ResNet-18 model, future work will validate the proposed technique across additional model architectures, application domains, and heterogeneous edge platforms with specialized accelerators. To achieve real-world applicability, dynamic batch-size selection based on real-time system profiles under concurrent workloads should be explored. Furthermore, integrating advanced optimization techniques, such as the momentum-based Adam optimizer or other adaptive learning rate methods, to enhance training stability should be investigated. However, integrating such sophisticated optimization techniques could introduce additional computational overhead compared to standard stochastic gradient descent in resource-constrained settings.

\bibliographystyle{IEEEtran}
\bibliography{references}
\end{document}